\documentclass{article}
\usepackage{spconf,amsmath,epsfig}
\usepackage[bookmarks=false]{hyperref}
\usepackage{url}            
\usepackage{booktabs}
\usepackage{amssymb}
\usepackage{amsmath}
\usepackage{algorithm}
\usepackage{algorithmic}
\usepackage{multirow}
\usepackage{listings}
\usepackage{enumitem}
\usepackage{subfigure}
\usepackage{marvosym}
\usepackage{amsthm}

\usepackage{mathrsfs}
\usepackage{graphicx}  
\begin{document}\sloppy

\def\x{{\mathbf x}}
\def\L{{\cal L}}

\title{Easy Transfer Learning By Exploiting Intra-domain Structures}
%
\name{Jindong Wang\textsuperscript{1}, Yiqiang Chen\textsuperscript{1,$\star$}\thanks{Corresponding author: Y. Chen. J. Wang and Y. Chen are also affiliated with University of Chinese Academy of Sciences, Beijing 100190, China. 978-1-5386-1737-3/18/\$31.00 ©2019 IEEE.}, Han Yu\textsuperscript{2}, Meiyu Huang\textsuperscript{3}, Qiang Yang\textsuperscript{4}}
\address{\textsuperscript{1}Beijing Key Lab. of Mobile Computing and Pervasive Device, Inst. of Computing Tech., CAS\\
	\textsuperscript{2}School of Computer Science and Engineering, Nanyang Technological University, Singapore\\
	\textsuperscript{3}Qian Xuesen Lab. of Space Tech., China Academy of Space Tech., Beijing, China\\
	\textsuperscript{4}Hong Kong University of Science and Technology, Hong Kong, China\\
	\{wangjindong, yqchen\}@ict.ac.cn, han.yu@ntu.edu.sg, huangmeiyu@qxslab.cn, qyang@cse.ust.hk}

\maketitle

\begin{abstract}
Transfer learning aims at transferring knowledge from a well-labeled domain to a similar but different domain with limited or no labels. Unfortunately, existing learning-based methods often involve intensive model selection and hyperparameter tuning to obtain good results. Moreover, cross-validation is not possible for tuning hyperparameters since there are often no labels in the target domain. This would restrict wide applicability of transfer learning especially in computationally-constraint devices such as wearables. In this paper, we propose a practically \textit{Easy Transfer Learning (EasyTL)} approach which requires no model selection and hyperparameter tuning, while achieving competitive performance. By exploiting intra-domain structures, EasyTL is able to learn both non-parametric transfer features and classifiers. Extensive experiments demonstrate that, compared to state-of-the-art traditional and deep methods, EasyTL satisfies the \textit{Occam's Razor} principle: it is extremely easy to implement and use while achieving comparable or better performance in classification accuracy and much better computational efficiency. Additionally, it is shown that EasyTL can increase the performance of existing transfer feature learning methods.
\end{abstract}
\section{Introduction}
The success of multimedia applications depends on the availability of sufficient labeled data to train machine learning models. However, it is often expensive and time-consuming to acquire massive amounts of labeled data. Transfer learning (TL), or domain adaptation~\cite{pan2010survey} is a promising strategy to enhance the learning performance on a target domain with few or none labels by leveraging knowledge from a well-labeled source domain. Since the source and target domains have different distributions, numerous methods have been proposed to reduce the distribution divergence~\cite{wei2018learning,wang2018visual,wang2018stratified,zhang2017joint,gong2012geodesic}.

\begin{figure}[t!]
	\centering
	\includegraphics[scale=0.4]{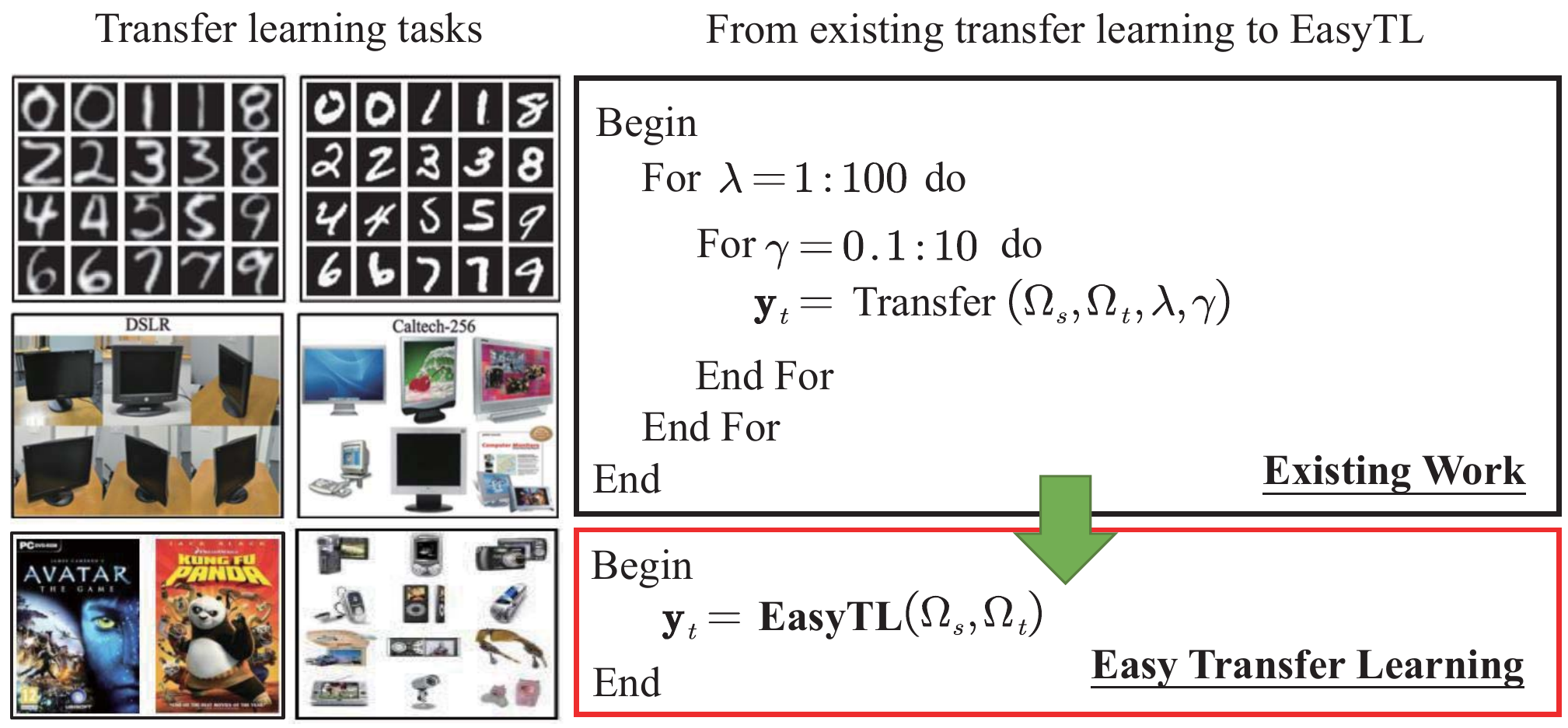}
	\vspace{-.15in}
	\caption{A brief illustration of EasyTL}
	\label{fig-intro}
	\vspace{-.2in}
\end{figure}

\begin{table}[t!]
	\caption{Comparison between EasyTL and recent TL methods on Office-Home~\cite{venkateswara2017deep} dataset}

	\vspace{.1in}
	\label{tb-comp}
	\resizebox{.48\textwidth}{!}{
		\begin{tabular}{|c|c|c|c|c|}
			\hline
			Method & MEDA~\cite{wang2018visual} & DANN~\cite{ganin2015unsupervised} & CDAN~\cite{long2018conditional} & EasyTL \\ \hline
			Accuracy & 60.3 & 57.6 & 62.8 & \textbf{63.3} \\ \hline
			Hyperparameter & $\lambda,\eta,p$ & $\lambda$ & $\lambda$ & \textbf{None} \\ \hline
			Network & \multicolumn{4}{c|}{ResNet50} \\ \hline
	\end{tabular}}
	\vspace{-.15in}
\end{table}

Unfortunately, despite the great success achieved by existing TL methods, it is notoriously challenging to apply them to a real situation since we cannot determine the best TL model and their optimal hyperparameters. The reasons are three-fold. Firstly, most traditional and deep TL methods are \textit{parametric} methods~\cite{wang2018visual,wei2018learning,zhang2017joint,ganin2015unsupervised} that have to go through an intensively expensive and time-consuming process to tune a lot of hyperparameters~(e.g. Fig.~\ref{fig-intro}; MEDA, DANN, and CDAN in Table~\ref{tb-comp}). Secondly, \textit{cross-validation}, which is the most common strategy to select models and tune hyperparameters, is \textit{not available} in TL since there are often no labeled data in the target domain~\cite{pan2010survey}. Thirdly, although the recent AutoML methods can automatically tune the hyperparameters via tree pruning, boosting, or neural architecture search~\cite{quanming2018taking}, they are unable to handle the different distributions between domains in TL and typically take a long time to converge. 

These challenges seriously restrict the real application of TL, especially on small devices that require instant local computing with limited resources such as wearables. Is it possible to develop an easy but powerful TL algorithm to circumvent model selection and parameter tuning, but have competitive performance, i.e. satisfying \textbf{Occam's Razor} principle~\cite{rasmussen2001occam}?

In this paper, we make the \textit{first} attempt towards addressing this challenge by proposing a practically \textbf{Easy Transfer Learning (EasyTL)} approach. EasyTL is able to perform knowledge transfer across domains \textit{without} the need for model selection and hyperparameter tuning~(Table~\ref{tb-comp}). By exploiting \textit{intra-domain stuctures}, EasyTL learns both non-parametric transfer features by \textit{intra-domain alignment} and transfer classifier by \textit{intra-domain programming}. Thus, it is able to avoid negative transfer~\cite{pan2010survey}. Furthermore, EasyTL can also \textit{increase} the performance of existing TL methods by serving as their final classifier via intra-domain programming.

We conduct extensive experiments on public TL datasets. Both visual domain adaptation and cross-domain sentiment analysis demonstrated significant superiority of EasyTL in classification accuracy and computational efficiency over state-of-the-art traditional and deep TL methods. In short, EasyTL has the following characteristics:

\begin{itemize}[noitemsep,nolistsep]
	\item \textit{Easy:}~EasyTL is extremely easy to implement and use. Therefore, it eliminates the need for model selection or hyperparameter tuning in transfer learning.
	\item \textit{Accurate:}~EasyTL produces competitive results in several popular TL tasks compared to state-of-the-art traditional and deep methods.
	\item \textit{Efficient:}~EasyTL is significantly more efficient than other methods. This makes EasyTL more suitable for resource-constrained devices such as wearables.
	\item \textit{Extensible:}~EasyTL can increase the performance of existing TL methods by replacing their classifier with intra-domain programming.
\end{itemize}

\section{Related Work}
EasyTL significantly differs from existing work in the following three aspects:

\textit{Transfer learning.} Existing TL methods can be summarized into two main categories: (a)~\textit{instance reweighting}, which reuses samples from the source domain according to some weighting technique; and (b)~\textit{feature transformation}, which performs subspace learning or distribution adaptation~\cite{wei2018learning,wang2018visual,pan2011domain,sun2016return,wang2017balanced}. Unfortunately, these methods are all \textit{parametric} approaches. They depend on extensive hyperparameter tuning through cross-validation for the feature transformation~\cite{pan2011domain,wei2018learning}, or the prediction model~\cite{gong2012geodesic,sun2016return}, or both~\cite{wang2017balanced,wang2018visual}. Most methods require multiple iterations of training~\cite{wang2017balanced,wei2018learning}. The L2T framework~\cite{wei2017learning} is similar to EasyTL in spirit, but L2T is still based on model iteration and parameter tuning. Deep TL methods~\cite{long2018conditional,long2017deep,venkateswara2017deep,ganin2015unsupervised} require heavy hyperparameter tuning. In TL, \textit{cross-validation} is often not available since there are almost no labeled data in the target domain~\cite{pan2010survey}. In contrast, EasyTL is a non-parametric TL approach that directly learns from \textit{intra-domain structures}, which requires no model selection and hyperparameter tuning and much more efficient than existing methods.

\textit{Non-parametric learning.} Nearest-neighbor (NN) classifier is the most common non-parametric method. However, NN computes the distance between each sample in two domains, which is more likely to be influenced by domain shift. Naive Bayes NN (NBNN) classifier is used for domain adaptation~\cite{tommasi2013frustratingly}, which still requires hyperparameter tuning and iterations. Nearest Centroid (NC) classifier is based on the distance between each target sample to the class center of the source domain. EasyTL uses a linear programming to get the softmax probabilities (float weights), while NC is basically 0/1 weights. This means that EasyTL could not only consider the relationship between sample and center, but also the relations of other samples. Open set DA~\cite{panareda2017open} has similar idea with EasyTL, while it solves a binary programming and requires other classifiers.

\textit{Automated machine learning.} Recent years have witnessed the advance of Automated Machine Learning~(AutoML)~\cite{quanming2018taking}. AutoML produces results without model selection and parameter tuning. However, no existing AutoML frameworks can handle TL tasks since they assume the training and test data are in the same distribution. AutoDIAL~\cite{maria2017autodial} is similar to EasyTL, which includes automatic domain alignment layers for deep networks. However, AutoDIAL still requires many parameters in the network to be tuned.

\section{Easy Transfer Learning}
\label{sec-method}

We follow the most common TL settings in existing work~\cite{wang2018visual,wei2018learning,zhang2017joint}. We are given a labeled source domain $\Omega_s=\{(\mathbf{x}^s_{i},y^s_{i})\}^{n_s}_{i=1}$ and an unlabeled target domain $\Omega_t=\{\mathbf{x}^t_{j}\}^{n_t}_{j=1}$. The feature space $\mathcal{X}_s = \mathcal{X}_t$, label space $\mathcal{Y}_s = \mathcal{Y}_t$, but marginal distribution $P_s(\mathbf{x}_s) \ne P_t(\mathbf{x}_t)$ with conditional distribution $Q_s(y_s|\mathbf{x}_s) \ne Q_t(y_t|\mathbf{x}_t)$. The goal is to predict the labels $\mathbf{y}_t \in \mathcal{Y}_t$ for the target domain.

\subsection{Motivation}

It is challenging to design a TL method that requires no model selection and hyperparameter tuning. On one hand, a simple NN (Nearest Neighbor) classifier may suffice, while NN suffers in handling the distribution divergence between domains. On the other hand, existing TL methods such as GFK~\cite{gong2012geodesic} and BDA~\cite{wang2017balanced} could reduce the distribution divergence, while they require tuning a lot of hyperparameters. Combining both of them without considering domain structures may easily result in \textit{negative transfer}~\cite{pan2010survey}, which dramatically hurts the performance of TL. Currently, there is no such effort.

\begin{figure}[t!]
	\centering
	\includegraphics[scale=.26]{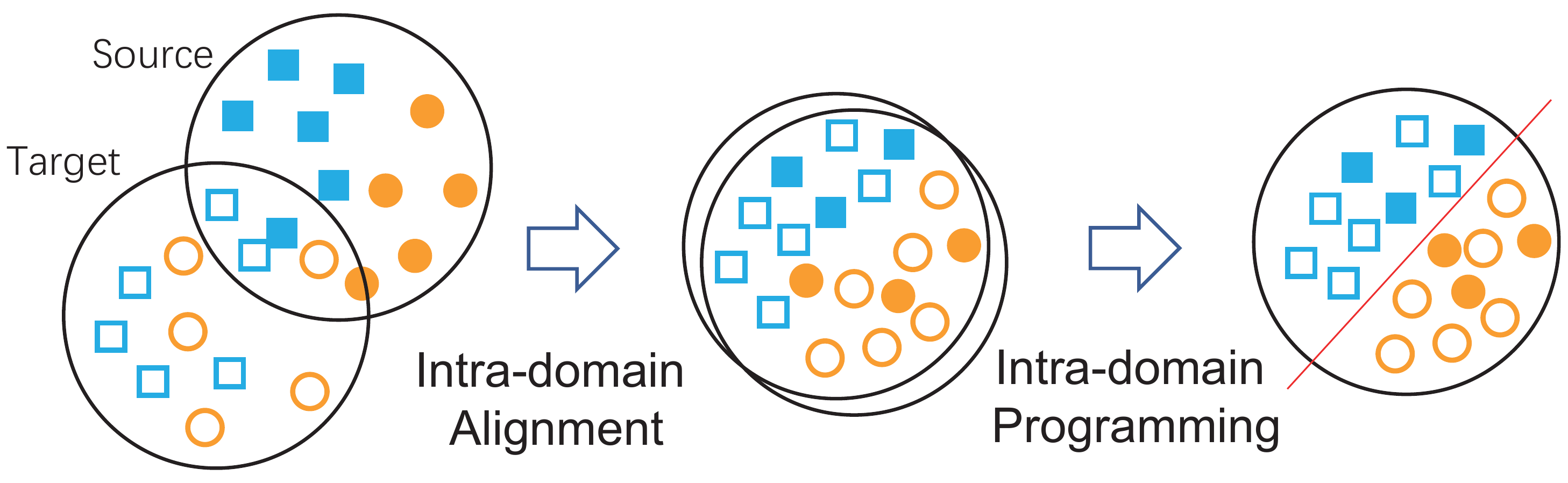}
	\vspace{-.1in}
	\caption{The procedure of EasyTL. The colored boxes and circles denote samples from the source and target domains. The red line denotes the classifier.}
	\label{fig-main}
	\vspace{-.15in}
\end{figure}

In this work, we propose a practically \textit{Easy Transfer Learning~(EasyTL)} approach to learn non-parametric transfer features and classifier with competitive performance to existing \textit{heavy} methods. In short, to be consistent with \textit{Occam's Razor} principle~\cite{rasmussen2001occam}. In light of the recent advance in representation learning~\cite{hjelm2018learning}, incorporating knowledge of locality can greatly improve the representation’s quality. Therefore, Instead of learning sample-wise distance, EasyTL focuses on exploiting the \textit{intra-domain structure}. The main part of EasyTL is a novel non-parametric \textit{Intra-domain programming} classifier, while remains open for adopting existing methods for \textit{Intra-domain alignment}. Intra-domain programming is able to learn discriminative transfer information from domains. Fig.~\ref{fig-main} shows the procedure of EasyTL. In the following sections, we first introduce the proposed intra-domain programming. Then, we show how to adapt a non-parametric feature learning method for intra-domain alignment.

\subsection{Intra-domain Programming}

\begin{figure}[t!]
	\centering
	\setlength{\fboxrule}{1pt} 
	\setlength{\fboxsep}{0.05cm}
	\fbox{\includegraphics[scale=0.36]{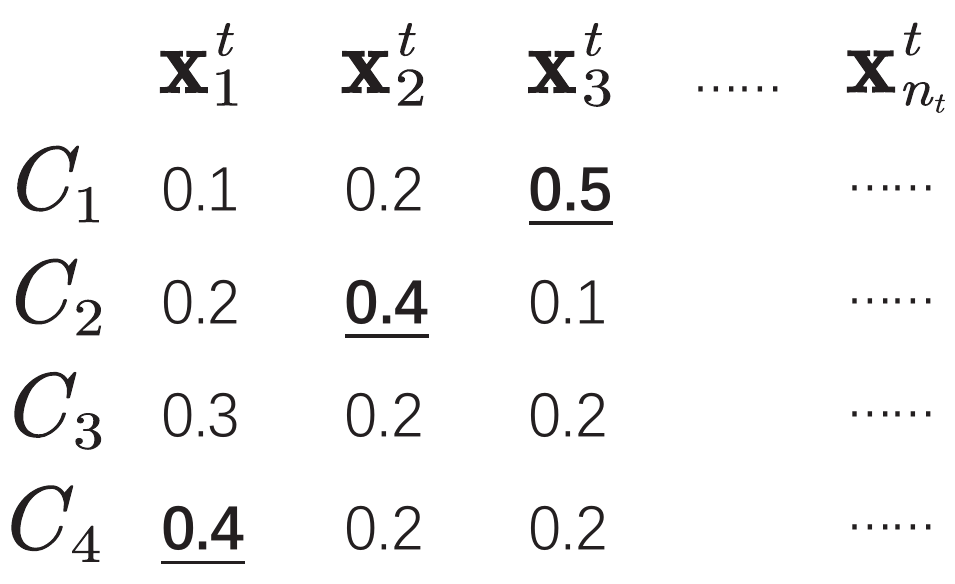}} 
	\vspace{-.1in}
	\caption{An example of the probability annotation matrix.}
	\label{fig-lp}
	\vspace{-.15in}
\end{figure}

Intra-domain programming directly learns a transfer classifier for TL problem and provides reliable likelihood information for intra-domain alignment. We introduce the concept of the \textit{Probability Annotation Matrix}, based on which we can build the non-parametric transfer classifier. Formally, let $c \in \{1,\cdots,C\}$ denote the class label, a matrix $\mathbf{M} \in \mathbb{R}^{C \times n_t}$ with its element $0 \le M_{cj} \le 1$ is a probability annotation matrix. Here, the entry value $M_{cj}$ of $\mathbf{M}$ indicates the annotation probability of $\mathbf{x}^t_j$ belonging to class $c$. Fig.~\ref{fig-lp} illustrates the main idea of $\mathbf{M}$. There are 4 classes and $n_t$ target samples with example activation values. Similar to the widely-adopted softmax classifier in neural networks, the highest probability value for $\mathbf{x}^t_1$ is $\mathrm{0.4}$, which indicates that it belongs to $C_4$. The same goes for $\mathbf{x}^t_2$ and $\mathbf{x}^t_3$, etc. The real probability annotation values are to be learned by our proposed approach.

Instead of learning $\mathbf{y}^t$ directly, the algorithm focuses on learning the probability annotation matrix $\mathbf{M}$. In this way, the cost function can be formalized as:
\begin{equation}
\label{eq-cost}
\mathcal{J} = \sum_{j}^{n_t} \sum_{c}^{C} D_{cj} M_{cj},
\end{equation}
where the distance value $D_{cj}$ is an entry in a distance matrix $\mathbf{D}$. $D_{cj}$ denotes the distance between $\mathbf{x}^t_j$ and the $c$-th class center of the source domain $\Omega^{(c)}_s$. 

We denote $\mathbf{h}_c$ as the $c$-th class center of $\Omega^{(c)}_s$. Then, $D_{cj}$ can be computed by the Euclidean distance:
\begin{equation}
\label{eq-kernel}
D_{cj} = ||\mathbf{x}^t_{j} - \mathbf{h}_c||^2.
\end{equation}

The class center $\mathbf{h}_c$ on $\Omega_s$ can be calculated as:
\begin{equation}
\mathbf{h}_c=\frac{1}{|\Omega^{(c)}_s|} \sum_{i}^{n_s} \mathbf{x}^s_i \cdot \mathbb{I}(y^s_i = c),
\end{equation}
where $\mathbb{I}(\cdot)$ is an indicator function which evaluates to 1 if the condition is true, and 0 otherwise.

Consider the constraints to minimize the cost function in Eq.~(\ref{eq-cost}).
Firstly, note that the value of $M_{cj}$ is a probability measuring the confidence of $\mathbf{x}^t_j$ belonging to class $c$, such that the sum of the probability of one particular sample $\mathbf{x}^t_j$ belonging to all existing classes is $\mathrm{1}$. This is ensured by the following constraint:
\begin{equation}
\label{eq-constraint1}
\sum_{c}^{C} M_{cj} = 1, \forall j \in \{1,\cdots,n_t\}.
\end{equation}
Secondly, since $\Omega_s$ and $\Omega_t$ have the same label space (i.e. $\mathcal{Y}_s = \mathcal{Y}_t$), there must be at least one sample for any given class $c$. This is ensured by the following constraint:
\begin{equation}
\label{eq-con2}
M_{c j} = \mathop{\arg\max}_{r} M_{r j}, r \in \{1, \cdots,C\}, \forall c \in \{1, \cdots, C\}, \exists j.
\end{equation}

In fact, the ideal state of $M_{cj}$ should be a binary value (0 or 1), i.e. $M_{cj}=1$ iff $\mathbf{x}_j$ belongs to class $c$, otherwise $M_{cj}=0$. Therefore, we use the following formulation to replace Eq.~(\ref{eq-con2}) in the computation without affecting the results:
\begin{equation}
\label{eq-constraint2}
\sum_{j}^{n_t} M_{cj} \ge 1, \forall c \in \{1, \cdots, C\}.
\end{equation}

\textit{Learning objective.} Combining the cost function in Eq.~(\ref{eq-cost}) and the constraints in Eq.~(\ref{eq-constraint1}) and Eq.~(\ref{eq-constraint2}), the final learning objective becomes:
\begin{equation}
\label{eq-final}
\begin{split}
\min \quad &\mathcal{J} = \sum_{j}^{n_t} \sum_{c}^{C} D_{cj} M_{cj}. \\
s.t. \quad &\begin{cases}
0 \le M_{cj} \le 1\\
\sum_{c}^{C} M_{cj} = 1, \forall j \in \{1,\cdots,n_t\} \\
\sum_{j}^{n_t} M_{cj} \ge 1, \forall c \in \{1, \cdots, C\}
\end{cases}
\end{split}
\end{equation}
Solving Eq.~(\ref{eq-final}) requires us to solve a linear programming~(LP) problem. There are existing solution packages such as PuLP~\footnote{\url{https://pypi.python.org/pypi/PuLP/1.1}} for solving the above linear programming problem efficiently. Then, $\mathbf{M}$ can be obtained. Eventually, the label of $\mathbf{x}^t_j$ is given by the softmax function:
\begin{equation}
\label{eq-softmax}
y^t_j = \mathop{\arg\max}_{r} \ \ \frac{M_{rj}}{\sum_{c}^{C} M_{cj}} \quad \text{for } r \in \{1,\cdots,C\}.
\end{equation}
It is noticeable that this classifier does \textbf{not} involve any parameters to tune explicitly. This is significantly different from well-established classifiers such as SVM that needs to tune numerous hyperparameters. In fact, intra-domain programming can be used alone for TL problems.

\subsection{Intra-domain Alignment} 

Intra-domain alignment serves as the transfer feature learning methods for EasyTL. Although EasyTL can be used directly in TL with good performance, it can also be combined with transfer feature learning to eliminate feature distortions in the original space. On the other hand, existing transfer feature learning methods can also be extended using the previous \textit{intra-domain programming} to enhance their performance.

Recall our inspiration that learning the structure of locality will help learn useful representations~\cite{hjelm2018learning}. Therefore, if we perform \textit{personalized} transfer feature learning within each subspace, we could further reduce the domain divergence. For simplicity and efficiency, we only perform feature learning from the source domain to the subspace of the target domain.

Inspired by the non-parametric feature learning method CORAL~\cite{sun2016return}, the intra-programming process of EasyTL can be formulated as:
\begin{equation}
\label{eq-coral-feature}
\mathbf{z}^r = \begin{cases}
\mathbf{x}^r \cdot (cov(\mathbf{x}^s) + \mathbf{E}_s)^{-\frac{1}{2}} \cdot (cov(\mathbf{x}^t) + \mathbf{E}_t)^{\frac{1}{2}} & \text{if }r = s \\
\mathbf{x}^r & \text{if } r = t
\end{cases}
\end{equation}
where $cov(\cdot)$ is the covariance matrix. $\mathbf{E}_s$ and $\mathbf{E}_t$ are identity matrices with equal sizes to $\Omega_s$ and $\Omega_t$, respectively. We can treat this step as a \textit{re-coloring} process of each subspace~\cite{sun2016return}. Eq.~(\ref{eq-coral-feature}) aligns the two distributions by re-coloring whitened source features with the covariance of target distributions.

\textit{Remark:} Other than CORAL, EasyTL can also choose other popular methods for feature learning such as BDA~\cite{wang2017balanced} and GFK~\cite{gong2012geodesic}. We choose CORAL for its computational efficiency and that it contains no other parameters to tune.

\subsection{Computational Complexity} 

We use the big-$O$ notation to analyze the complexity of EasyTL. Intra-domain alignment takes at most $O(C n_s^3)$. The complexity of intra-domain programming is $O(n_t^3 C^3)$ given that it is a linear programming problem (Eq.~(\ref{eq-final}))~\cite{megiddo1986complexity}. EasyTL can be made more efficient with the low-rank representations and other fast computing algorithms.

As an intuitive comparison of classifiers, we compare intra-domain programming with the well-established SVM. A \textit{single} round of SVM training and prediction takes $O((n_s + n_t)^3)$ computations~\cite{abdiansah2005svm}. This implies that the time complexity of intra-domain programming is comparable to a single round of SVM. However, SVM still needs a couple of rounds for hyperparameter tuning (e.g. \textit{kerneltype}, \textit{constraints}) before getting optimal performance. Therefore, EasyTL is theoretically more efficient than SVM. We will experimentally show the efficiency of EasyTL in the next sections.

\begin{algorithm}[t!]
	\caption{EasyTL: Easy Transfer Learning}  
	\label{algo-auto}  
	\renewcommand{\algorithmicrequire}{\textbf{Input:}} 
	\renewcommand{\algorithmicensure}{\textbf{Output:}}
	\begin{algorithmic}[1]  
		\REQUIRE 
		Feature matrix $\mathbf{x}_s,\mathbf{x}_t$ for $\Omega_s$ and $\Omega_t$, respectively; and label vector $\mathbf{y}_s$ for $\Omega_s$\\
		\ENSURE 
		Predicted label vector $\mathbf{y}_t$ for target domain.\\
		\STATE (Optional) Perform intra-domain alignment via Eq.~(\ref{eq-coral-feature})
		\STATE Solve Eq.~(\ref{eq-final}) to obtain the probability annotation matrix $\mathbf{M}$ and compute $\mathbf{y}_t$ using Eq.~(\ref{eq-softmax})
		\RETURN Label vector $\mathbf{y}_t$ for $\Omega_t$ 
	\end{algorithmic}
\end{algorithm}

\begin{table*}[t!]
	\centering
	\caption{Accuracy~(\%) on Image-CLEF DA and Office-Home datasets using ResNet features}
	\label{tb-image}
	\resizebox{.78\textwidth}{!}{
		\begin{tabular}{|c|c|c|c|c|c|c|c|c|c|c|c|c|c|}
			\hline
			ID & Task & ResNet & 1NN & SVM & TCA & GFK & BDA & CORAL & DANN & JAN & CDAN & EasyTL(c) & EasyTL \\ \hline \hline
			1 & C $\rightarrow$ I & 78.0 & 83.5 & 86.0 & 89.3 & 86.3 & 90.8 & 83.0 & 87.0 & 89.5 & 91.2 & 85.5 & \textbf{91.5} \\ \hline
			2 & C $\rightarrow$ P & 65.5 & 71.3 & 73.2 & 74.5 & 73.3 & 73.7 & 71.5 & 74.3 & 74.2 & 77.2 & 72.0 & \textbf{77.7} \\ \hline
			3 & I $\rightarrow$ C & 91.5 & 89.0 & 91.2 & 93.2 & 93.0 & 94.0 & 88.7 & 96.2 & 94.7 & 96.7 & 93.3 & \textbf{96.0} \\ \hline
			4 & I $\rightarrow$ P & 74.8 & 74.8 & 76.8 & 77.5 & 75.5 & 75.3 & 73.7 & 75.0 & 76.8 & 78.3 & 78.5 & \textbf{78.7} \\ \hline
			5 & P $\rightarrow$ C & 91.2 & 76.2 & 85.8 & 83.7 & 82.3 & 83.5 & 72.0 & 91.5 & 91.7 & 93.7 & 91.0 & \textbf{95.0} \\ \hline
			6 & P $\rightarrow$ I & 83.9 & 74.0 & 80.2 & 80.8 & 78.0 & 77.8 & 71.3 & 86.0 & 88.0 & \textbf{91.2} & 89.5 & 90.3 \\ \hline
			- & AVG & 80.7 & 78.1 & 82.2 & 83.2 & 81.4 & 82.5 & 76.7 & 85.0 & 85.8 & 88.1 & 85.0 & \textbf{88.2} \\ \hline \hline
			7 & Ar $\rightarrow$ Cl & 34.9 & 45.3 & 45.3 & 38.3 & 38.9 & 38.9 & 42.2 & 45.6 & 45.9 & 46.6 & 51.6 & \textbf{52.8} \\ \hline
			8 & Ar $\rightarrow$ Pr & 50.0 & 60.1 & 65.4 & 58.7 & 57.1 & 54.8 & 59.1 & 59.3 & 61.2 & 65.9 & 68.1 & \textbf{72.1} \\ \hline
			9 & Ar $\rightarrow$ Rw & 58.0 & 65.8 & 73.1 & 61.7 & 60.1 & 58.2 & 64.9 & 70.1 & 68.9 & 73.4 & 74.2 & \textbf{75.9} \\ \hline
			10 & Cl $\rightarrow$ Ar & 37.4 & 45.7 & 43.6 & 39.3 & 38.7 & 36.2 & 46.4 & 47.0 & 50.4 & 55.7 & 53.1 & \textbf{55.0} \\ \hline
			11 & Cl $\rightarrow$ Pr & 41.9 & 57.0 & 57.3 & 52.4 & 53.1 & 53.1 & 56.3 & 58.5 & 59.7 & 62.7 & 62.9 & \textbf{65.9} \\ \hline
			12 & Cl $\rightarrow$ Rw & 46.2 & 58.7 & 60.2 & 56.0 & 55.5 & 50.2 & 58.3 & 60.9 & 61.0 & 64.2 & 65.3 & \textbf{67.6} \\ \hline
			13 & Pr $\rightarrow$ Ar & 38.5 & 48.1 & 46.8 & 42.6 & 42.2 & 42.1 & 45.4 & 46.1 & 45.8 & 51.8 & 52.8 & \textbf{54.4} \\ \hline
			14 & Pr $\rightarrow$ Cl & 31.2 & 42.9 & 39.1 & 37.5 & 37.6 & 38.2 & 41.2 & 43.7 & 43.4 & \textbf{49.1} & 45.8 & 46.9 \\ \hline
			15 & Pr $\rightarrow$ Rw & 60.4 & 68.9 & 69.2 & 64.1 & 64.6 & 63.1 & 68.5 & 68.5 & 70.3 & 74.5 & 73.5 & \textbf{74.7} \\ \hline
			16 & Rw $\rightarrow$ Ar & 53.9 & 60.8 & 61.1 & 52.6 & 53.8 & 50.2 & 60.1 & 63.2 & 63.9 & \textbf{68.2} & 62.2 & 63.8 \\ \hline
			17 & Rw $\rightarrow$ Cl & 41.2 & 48.3 & 45.6 & 41.7 & 42.3 & 44.0 & 48.2 & 51.8 & 52.4 & \textbf{56.9} & 50.2 & 52.3 \\ \hline
			18 & Rw $\rightarrow$ Pr & 59.9 & 74.7 & 75.9 & 70.5 & 70.6 & 68.2 & 73.1 & 76.8 & 76.8 & \textbf{80.7} & 76.0 & 78.0 \\ \hline
			- & AVG & 46.1 & 56.4 & 56.9 & 51.3 & 51.2 & 49.8 & 55.3 & 57.6 & 58.3 & 62.8 & 61.3 & \textbf{63.3} \\ \hline \hline
			- & Avg rank~\cite{demvsar2006statistical} & 12 & 7 & 6 & 9 & 10 & 11 & 8 & 5 & 4 & 2 & \underline{3} & \textbf{1} \\ \hline
			
	\end{tabular}}
\end{table*}

\begin{table}[t!]
	\vspace{-.2in}
	\caption{Accuracy~(\%) on Amazon Review dataset}
	\label{tb-amazon}
	\resizebox{.5\textwidth}{!}{
		\begin{tabular}{|c|c|c|c|c|c|c|c|c|c|}
			\hline
			Method & 1NN & TCA & GFK & SA & BDA & CORAL & JGSA & EasyTL(c) & EasyTL \\ \hline \hline
			B $\rightarrow$ D & 49.6 & 63.6 & 66.4 & 67.0 & 64.2 & 71.6 & 66.6 & 78.4 & \textbf{79.8} \\ \hline
			B $\rightarrow$ E & 49.8 & 60.9 & 65.5 & 70.8 & 62.1 & 65.1 & 75.0 & 77.5 & \textbf{79.7} \\ \hline
			B $\rightarrow$ K & 50.3 & 64.2 & 69.2 & 72.2 & 65.4 & 67.3 & 72.1 & 79.2 & \textbf{80.9} \\ \hline
			D $\rightarrow$ B & 53.3 & 63.3 & 66.3 & 67.5 & 62.4 & 70.1 & 55.5 & 79.5 & \textbf{79.9} \\ \hline
			D $\rightarrow$ E & 51.0 & 64.2 & 63.7 & 67.1 & 66.3 & 65.6 & 67.3 & 77.4 & \textbf{80.8} \\ \hline
			D $\rightarrow$ K & 53.1 & 69.1 & 67.7 & 69.4 & 68.9 & 67.1 & 65.6 & 80.4 & \textbf{82.0} \\ \hline
			E $\rightarrow$ B & 50.8 & 59.5 & 62.4 & 61.4 & 59.2 & 67.1 & 51.6 & 73.0 & \textbf{75.0} \\ \hline
			E $\rightarrow$ D & 50.9 & 62.1 & 63.4 & 64.9 & 61.6 & 66.2 & 50.8 & 73.1 & \textbf{75.3} \\ \hline
			E $\rightarrow$ K & 51.2 & 74.8 & 73.8 & 70.4 & 74.7 & 77.6 & 55.0 & 84.6 & \textbf{84.9} \\ \hline
			K $\rightarrow$ B & 52.2 & 64.1 & 65.5 & 64.4 & 62.7 & 68.2 & 58.3 & 75.2 & \textbf{76.5} \\ \hline
			K $\rightarrow$ D & 51.2 & 65.4 & 65.0 & 64.6 & 64.3 & 68.9 & 56.4 & 73.8 & \textbf{76.3} \\ \hline
			K $\rightarrow$ E & 52.3 & 74.5 & 73.0 & 68.2 & 74.0 & 75.4 & 51.7 & 82.0 & \textbf{82.5} \\ \hline \hline
			AVG & 51.3 & 65.5 & 66.8 & 67.3 & 65.5 & 69.1 & 60.5 & 77.8 & \textbf{79.5} \\ \hline
	\end{tabular}}
	\vspace{-.2in}
\end{table}

\section{Experimental Evaluation}
\label{sec-exp}


\subsection{Experimental Setup}
We use four popular TL datasets:~1)~\textbf{Amazon Review}~\cite{chen2012marginalized} is a cross-domain sentiment analysis dataset that contains positive and negative reviews of four kinds of products: Kitchen appliance~(K), DVDs~(D), Electronics~(E), and Books~(B). 2)~\textbf{Office-Caltech}~\cite{gong2012geodesic} contains 10 common classes of images in Amazon~(A), DSLR~(D), Webcam~(W), and Caltech~(C). 3)~\textbf{Image-CLEF DA}~\cite{long2017deep} contains 12 categories of images belonging to 3 domains: Caltech~(C), ImageNet~(I), and Pascal~(P). 4)~\textbf{Office-Home}~\cite{venkateswara2017deep} contains 15,500 images of 65 categories from 4 domains: Art~(Ar), Clipart~(Cl), Product~(Pr), and Real-world~(Rw). Within each dataset, any two domains can be source and target to construct TL tasks.

State-of-the-art traditional and deep comparison methods: \textbf{NN} (Nearest Neighbor), \textbf{SVM} with linear kernel~\cite{sun2016return}, \textbf{TCA} (Transfer Component Analysis)~\cite{pan2011domain}, \textbf{GFK} (Geodesic Flow Kernel)~\cite{gong2012geodesic}, \textbf{SA}~(Subspace Alignment) \cite{fernando2013unsupervised}, \textbf{CORAL} (CORrelation ALignment)~\cite{sun2016return}, \textbf{BDA} (Balanced Distribution Adaptation)~\cite{wang2017balanced}, \textbf{JGSA} (Joint Geometrical and Statistical Alignment)~\cite{zhang2017joint}, \textbf{D-GFK}~\cite{wei2018learning}, \textbf{ResNet50}, \textbf{DANN} (Domain-adversarial Neural Networks)~\cite{ganin2015unsupervised}, \textbf{JAN} (Joint Adaptation Networks)~\cite{long2017deep}, and \textbf{CDAN} (Conditional Adversarial Adaptation Networks)~\cite{long2018conditional}. The source code of EasyTL is available at \url{http://transferlearning.xyz/code/traditional/EasyTL}.

Specifically, we use \textit{EasyTL(c)} to denote intra-domain programming since it can serve as a TL method alone, while \textit{EasyTL} denotes the full method. By following the standard protocol in~\cite{sun2016return}, we adopt linear SVM for traditional TL methods. For the amazon dataset, we use the 400-dimensional features~\cite{features}. For image datasets, we use the 2048-dimensional ResNet50 finetuned features~\cite{features}. For Office-Caltech datasets, we adopt the SURF features~\cite{wei2018learning}. Deep TL methods are only used in image datasets whose results are obtained from existing work. Classification accuracy is acting as the evaluation metric~\cite{pan2011domain,gong2012geodesic,sun2016return}.

\subsection{Results: Accuracy and Efficiency}

The results on Amazon Review is shown in Table~\ref{tb-amazon}. Table~\ref{tb-image} shows the results on ImageCLEF DA~(IDs 1$\sim$6) and Office-Home~(IDs 7$\sim$18). Results on Office-Caltech are in the code page due to space limit. We also report the average rank~\cite{demvsar2006statistical}, parameter and running time~(train+test) of several methods in Table~\ref{tb-para}. The results demonstrate that EasyTL outperforms all comparison methods in both sentiment and image data. EasyTL(c) can also achieve competitive performance than most methods. Note that except 1NN and EasyTL, other methods all require hyperparameter tuning. It clearly indicates the superiority of EasyTL in accuracy and efficiency.

There are more insightful findings. 1)~On the larger and more challenging Office-Home dataset, the performance of EasyTL(c) is only slightly worse than CDAN, while it outperforms all other comparison methods. Compared to deep TL methods (DANN, JAN, and CDAN), although EasyTL takes finetuned features as inputs, it only requires \textit{one} distinct finetuning process while deep methods have to run the network multiple times to get optimal parameters. 2)~Moreover, in real applications, it is rather difficult and almost \textit{impossible} to search parameters since there are often little or none labeled data in the target domain. It seems that adversarial learning provides little contribution on TL compared to the power of ResNet in representation learning, and it is rather difficult to train an adversarial network. 3)~The results imply that the ``2-stage'' procedure (finetune+EasyTL) is better than ``1-stage'' methods (deep TL) even if they are end-to-end. All methods have their advantages and disadvantages. 4)~EasyTL tends to perform consistently well on rather balanced datasets~(Amazon Review and Image-CLEF DA). While on the unbalanced Office-Caltech dataset, the performance of EasyTL is limited. The research of improving EasyTL for unbalanced datasets remains as future research.

\begin{table}[t!]
	\vspace{-.2in}
	\caption{Average rank, parameter, and running time}
	\label{tb-para}
	\resizebox{.5\textwidth}{!}{
		\begin{tabular}{ccccccccc}
			\toprule
			Method  & TCA & GFK & CORAL & DANN & CDAN & EasyTL(c) & EasyTL \\ \hline
			Avg rank  & 9 & 10 & 8 & 4 & 2 & \underline{3} & \textbf{1} \\ 
			Parameter  & $d,\lambda,C,\gamma$ & $d,C,\gamma$ & $C,\gamma$ & $\lambda$ & $\lambda$ & \textbf{None} & \textbf{None} \\ 
			Time (s) & 119 & 127.4 & 126.5 & $>$1000 & $>$1000 & \textbf{23.8} & \underline{59.6} \\ \bottomrule
	\end{tabular}}
	\vspace{-.15in}
\end{table}

\subsection{Evaluation of Extensibility}

\begin{figure}[t!]
	\centering
	\vspace{-.1in}
	\subfigure[Image-CLEF DA]{
		\includegraphics[scale=0.29]{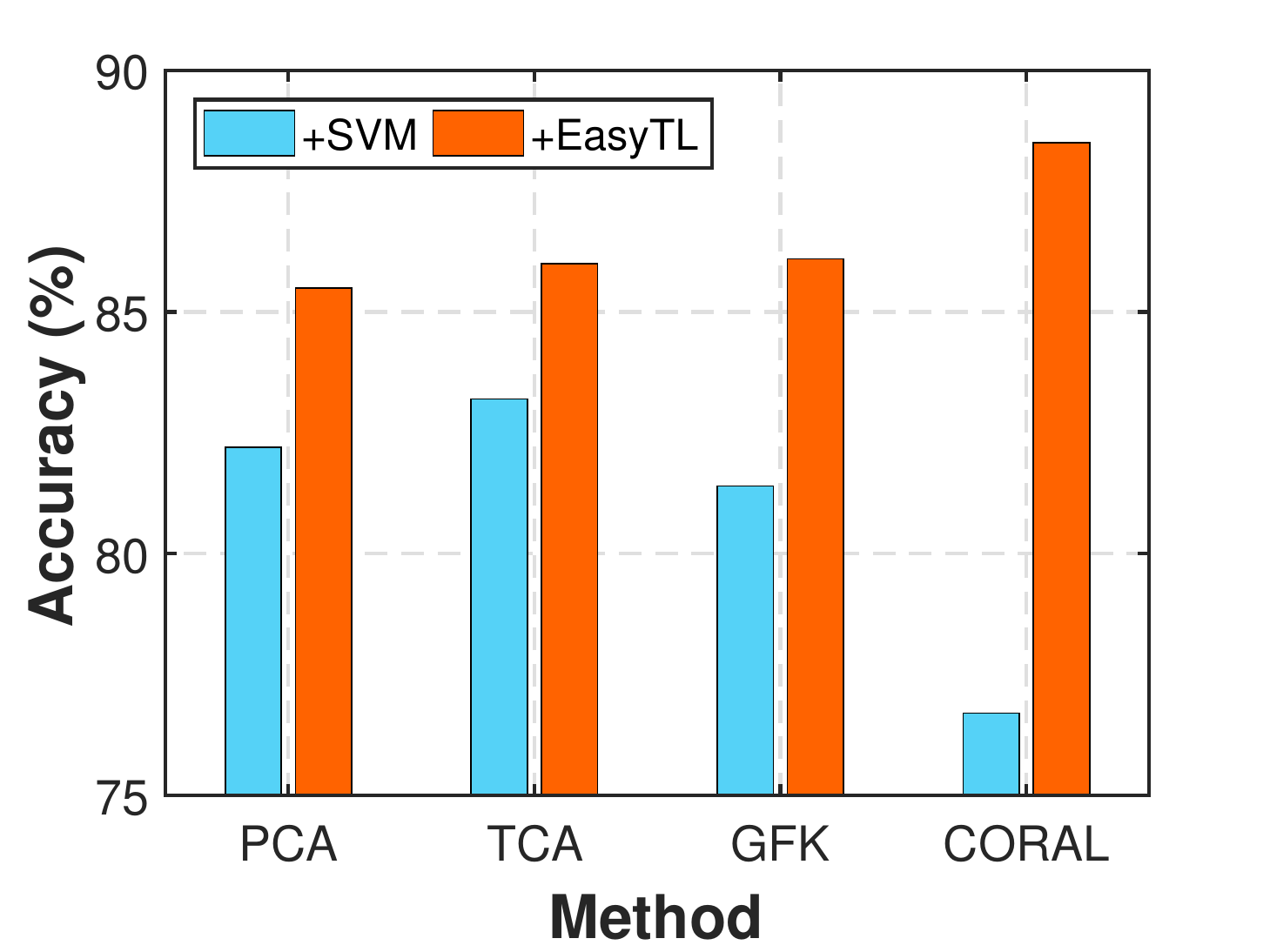}
		\label{fig-imageclef}}
	\hspace{-.2in}
	\subfigure[Office-Home]{
		\includegraphics[scale=0.29]{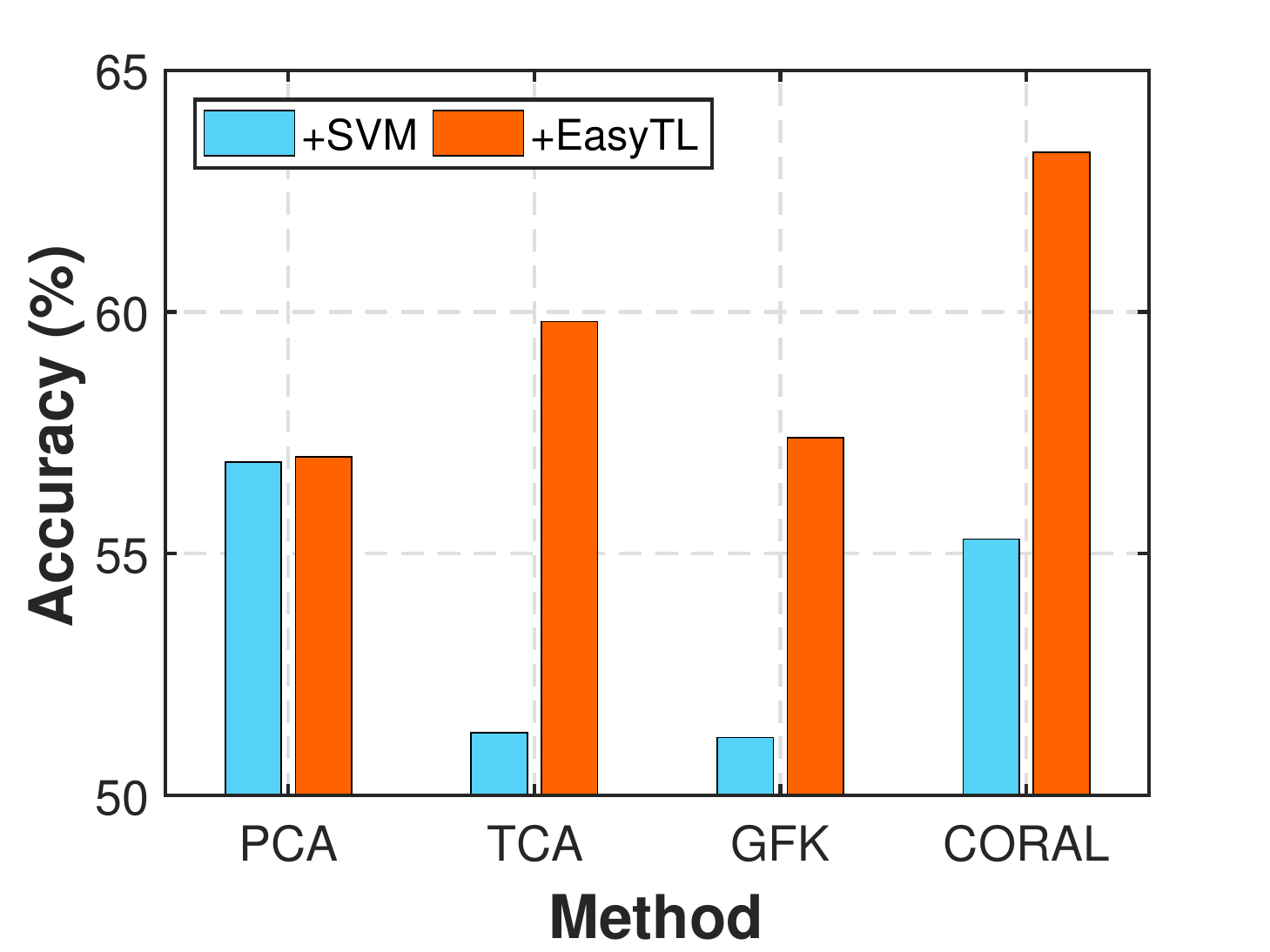}%
		\label{fig-vis-la}}
	\vspace{-.15in}
	\caption{Extending existing methods with EasyTL}
	\label{fig-extend}
	\vspace{-.1in}
\end{figure}

We evaluate the extensibility of EasyTL by extending existing TL methods with the classifier of EasyTL. We use PCA, TCA~\cite{pan2011domain}, GFK~\cite{gong2012geodesic}, and CORAL~\cite{sun2016return} for intra-domain alignment and compare the performance of +SVM and +EasyTL in Fig.~\ref{fig-extend}. The results show that: 1) By using feature learning methods other than CORAL, EasyTL can still achieve comparable performance to existing approaches. 2) More importantly, EasyTL does \textbf{not} rely on any particular feature learning methods to achieve good performance. 3) EasyTL can \textit{increase} the performances of existing TL methods. It clearly implies the extensibility of EasyTL in transfer learning.

\section{Conclusions and Future Work}
\label{sec-conclu}

In this paper, we propose EasyTL as the first non-parametric transfer learning approach by exploiting intra-domain structures features to learn non-parametric transfer features and classifiers. Thus, it requires no model selection and hyperparameter tuning. EasyTL is easy, accurate, efficient, and extensible, and can be applied directly in the resource-constrained devices such as wearables. Experiments demonstrate the superiority of EasyTL in accuracy and efficiency over several popular traditional and deep TL methods. EasyTL can also increase the performance of other TL methods.

EasyTL opens up some new research opportunities in TL:

\textit{Practical Transfer Learning.} Instead of pursuing good performance in accuracy, we do hope that EasyTL could open up a new door in research on practical transfer learning that focuses more on the actual usage of the methods: parameter tuning and model validation.


\textit{Automated Domain Selection.} Currently, EasyTL focuses on \textit{how to transfer}. In real applications, another important problem is \textit{what to transfer}. Some existing work focused on this issue, but they also require heavy parameter tuning. We will further explore this part by extending EasyTL.

\textit{Multi-label Transfer Learning.} EasyTL can be used in multi-label transfer learning by changing the value of probability annotation matrix. This makes EasyTL suitable for multi-label tasks. Existing multi-label transfer methods have to modify their classifiers. EasyTL can be more advantageous for these applications.

\section{Acknowledgments}

This work is supported by National Key R \& D Program of China
(2016YFB1001200), NSFC (61572471, 61702520), Hong Kong CERG projects (16209715, 16244616), and Nanyang Technological University, Nanyang Assistant Professorship (NAP).

\vspace{-.18cm}
\newcommand{\BIBdecl}{\setlength{\itemsep}{0.1 em}}
{
\small
\bibliographystyle{abbrv}
\bibliography{icme19}

\begin{thebibliography}{10}

\bibitem{abdiansah2005svm}
A.~Abdiansah and R.~Wardoyo.
\newblock Time complexity analysis of support vector machines in libsvm.
\newblock {\em J. Comput. Appl.}, 128(3):28--34, 2015.

\bibitem{chen2012marginalized}
M.~Chen, Z.~Xu, K.~Weinberger, and F.~Sha.
\newblock Marginalized denoising autoencoders for domain adaptation.
\newblock In {\em ICML}, 2012.

\bibitem{demvsar2006statistical}
J.~Dem{\v{s}}ar.
\newblock Statistical comparisons of classifiers over multiple data sets.
\newblock {\em JMLR}, 7(Jan):1--30, 2006.

\bibitem{fernando2013unsupervised}
B.~Fernando et~al.
\newblock Unsupervised visual domain adaptation using subspace alignment.
\newblock In {\em ICCV}, pages 2960--2967, 2013.

\bibitem{ganin2015unsupervised}
Y.~Ganin and V.~Lempitsky.
\newblock Unsupervised domain adaptation by backpropagation.
\newblock In {\em ICML}, pages 1180--1189, 2015.

\bibitem{gong2012geodesic}
B.~Gong, Y.~Shi, et~al.
\newblock Geodesic flow kernel for unsupervised domain adaptation.
\newblock In {\em CVPR}, pages 2066--2073, 2012.

\bibitem{hjelm2018learning}
R.~D. Hjelm et~al.
\newblock Learning deep representations by mutual information estimation and
  maximization.
\newblock In {\em ICLR}, 2019.

\bibitem{long2018conditional}
M.~Long, Z.~Cao, J.~Wang, et~al.
\newblock Conditional adversarial domain adaptation.
\newblock In {\em NeurIPS}, pages 1647--1657, 2018.

\bibitem{long2017deep}
M.~Long, J.~Wang, et~al.
\newblock Deep transfer learning with joint adaptation networks.
\newblock In {\em ICML}, pages 2208--2217, 2017.

\bibitem{maria2017autodial}
F.~Maria~Carlucci, L.~Porzi, et~al.
\newblock Autodial: Automatic domain alignment layers.
\newblock In {\em ICCV}, pages 5067--5075, 2017.

\bibitem{megiddo1986complexity}
N.~Megiddo.
\newblock {\em On the complexity of linear programming}.
\newblock 1986.

\bibitem{pan2010survey}
S.~J. Pan and Q.~Yang.
\newblock A survey on transfer learning.
\newblock {\em IEEE TKDE}, 22(10):1345--1359, 2010.

\bibitem{pan2011domain}
S.~J. Pan, Q.~Yang, et~al.
\newblock Domain adaptation via transfer component analysis.
\newblock {\em IEEE TNN}, 22(2):199--210, 2011.

\bibitem{panareda2017open}
P.~Panareda~Busto and J.~Gall.
\newblock Open set domain adaptation.
\newblock In {\em ICCV}, pages 754--763, 2017.

\bibitem{quanming2018taking}
Y.~Quanming et~al.
\newblock Taking human out of learning applications: A survey on automated
  machine learning.
\newblock {\em arXiv preprint:1810.13306}, 2018.

\bibitem{rasmussen2001occam}
C.~E. Rasmussen and Z.~Ghahramani.
\newblock Occam's razor.
\newblock In {\em NeurIPS}, pages 294--300, 2001.

\bibitem{sun2016return}
B.~Sun, J.~Feng, and K.~Saenko.
\newblock Return of frustratingly easy domain adaptation.
\newblock In {\em AAAI}, volume~6, page~8, 2016.

\bibitem{tommasi2013frustratingly}
T.~Tommasi and B.~Caputo.
\newblock Frustratingly easy nbnn domain adaptation.
\newblock In {\em ICCV}, pages 897--904, 2013.

\bibitem{venkateswara2017deep}
H.~Venkateswara et~al.
\newblock Deep hashing network for unsupervised domain adaptation.
\newblock In {\em CVPR}, pages 5018--5027, 2017.

\bibitem{wang2017balanced}
J.~Wang, Y.~Chen, S.~Hao, et~al.
\newblock Balanced distribution adaptation for transfer learning.
\newblock In {\em ICDM}, pages 1129--1134, 2017.

\bibitem{wang2018stratified}
J.~Wang, Y.~Chen, L.~Hu, et~al.
\newblock Stratified transfer learning for cross-domain activity recognition.
\newblock In {\em PerCom}, 2018.

\bibitem{features}
J.~Wang et~al.
\newblock Everything about transfer learning and domain adaptation.
\newblock \url{http://transferlearning.xyz}, 2018.

\bibitem{wang2018visual}
J.~Wang, W.~Feng, Y.~Chen, et~al.
\newblock Visual domain adaptation with manifold embedded distribution
  alignment.
\newblock In {\em ACM MM}, pages 402--410, 2018.

\bibitem{wei2018learning}
J.~Wei, J.~Liang, R.~He, and J.~Yang.
\newblock Learning discriminative geodesic flow kernel for unsupervised domain
  adaptation.
\newblock In {\em ICME}, pages 1--6, 2018.

\bibitem{wei2017learning}
Y.~Wei, Y.~Zhang, and Q.~Yang.
\newblock Transfer learning via learning to transfer.
\newblock In {\em ICML}, 2018.

\bibitem{zhang2017joint}
J.~Zhang, W.~Li, and P.~Ogunbona.
\newblock Joint geometrical and statistical alignment for visual domain
  adaptation.
\newblock In {\em CVPR}, 2017.

\end{thebibliography}
}

\end{document}